\documentclass{article}

\usepackage[T1]{fontenc}
\usepackage[utf8]{inputenc}
\usepackage{microtype}
\usepackage{graphicx}
\usepackage{booktabs}
\usepackage{xcolor}
\usepackage{hyperref}
\usepackage[preprint]{icml2025}

\usepackage{latexsym}
\usepackage{inconsolata}
\usepackage{amsmath,amssymb}
\usepackage{tabularx}
\usepackage{makecell}
\usepackage{siunitx}
\usepackage{enumitem}
\usepackage{capt-of}
\providecommand{\captionsetup}[1]{}
\usepackage{tcolorbox}
\tcbuselibrary{skins}

\newcommand{\draft}{\textsc{Draft}}
\newcommand{\verify}{\textsc{Verify}}
\newcommand{\scverify}{\textsc{SC-Verify}}
\newcommand{\cove}{\textsc{CoVe}}
\newcommand{\rarr}{\textsc{RARR}}
\newcommand{\method}{\textsc{CAR}}
\definecolor{cframe}{RGB}{120,160,205}
\definecolor{cbad}{RGB}{200,40,40}
\definecolor{cgood}{RGB}{40,140,70}
\definecolor{ckey}{RGB}{225,120,30}
\newcommand{\bad}[1]{\textcolor{cbad}{\textbf{#1}}}
\newcommand{\good}[1]{\textcolor{cgood}{\textbf{#1}}}
\newcommand{\key}[1]{\textcolor{ckey}{\textbf{#1}}}
\newtcolorbox{casebox}{enhanced,colback=white,colframe=cframe,
  boxrule=0.9pt,arc=3mm,left=5pt,right=5pt,top=5pt,bottom=5pt,
  boxsep=1pt,fontupper=\footnotesize}
\newcommand{\lab}[1]{\par\vspace{1.5pt}\noindent
  \tcbox[on line,colback=white,colframe=cframe,boxrule=0.6pt,
  arc=1.6mm,left=3pt,right=3pt,top=1pt,bottom=1pt,boxsep=0pt]
  {\footnotesize\textbf{#1}}\par\vspace{1pt}}
\newcommand{\promptbox}[1]{%
  \par\smallskip
  \noindent\fbox{%
    \begin{minipage}{\dimexpr\linewidth-2\fboxsep-2\fboxrule\relax}
      \small #1
    \end{minipage}%
  }%
  \par\smallskip
}

\icmltitlerunning{When Self-Verification Hurts}

\begin{document}

\twocolumn[
\icmltitle{When Self-Verification Hurts: \\ Calibrated Answer Repair for Financial Question Answering}

\icmlsetsymbol{equal}{*}
\begin{icmlauthorlist}
\icmlauthor{Celine Guo}{equal,nus}
\icmlauthor{Weida Liang}{equal,nus}
\icmlauthor{Yujian Fu}{nus}
\icmlauthor{Anqi Dang}{ntu}
\icmlauthor{Yanzheng Jin}{nus}
\icmlauthor{Kenji Kawaguchi}{nus}
\icmlauthor{Wenya Wang}{ntu}
\end{icmlauthorlist}

\icmlaffiliation{nus}{National University of Singapore, Singapore}
\icmlaffiliation{ntu}{Nanyang Technological University, Singapore}

\vskip 0.3in
]
\printAffiliationsAndNotice{\icmlEqualContribution}

\begin{abstract}
Financial question answering can fail even when a model extracts a plausible number from the supplied evidence: the answer may attach it to the wrong entity, period, unit, or accounting concept. Asking the model to verify and revise its draft can help, but open-ended revision may alter content that was already correct or inspect the wrong relation and reinforce the original error. We formulate repair as a decision that balances the information it may recover against the correct content it may damage. Based on this trade-off, we introduce \textbf{Calibrated Answer Repair} (\method{}), which first estimates whether a draft is likely to benefit from repair, then selects a targeted check, localizes the relevant evidence or draft span, and limits the permitted edit. Across FinRank, Fin-RATE, and FinQA with three open-weight models, \method{} outperforms generic verification in all dataset--model settings. At the system level, it recovers 14.6\% of initially missed answer requirements while damaging 2.1\% of initially correct ones. Its post-draft token cost is over 30 percent lower than the iterative revision alternatives. Ablations on two dataset--model settings show that evidence localization and bounded editing are the main contributors to \method{}'s gains.
\end{abstract}

\section{Introduction}
\label{sec:intro}

Financial reports rarely present an answer as an isolated number. Even a simple numerical question may require a system to combine narrative
text, tables, and intermediate calculations, and then determine which company,
reporting period, unit, and accounting concept a value describes
\citep{zhu2021tatqa,chen2021finqa,chen2022convfinqa}. Financial question
answering therefore requires both numerical reasoning and precise grounding in
the supplied evidence. Language models can succeed at finding or computing a
value while failing to ground it correctly: a number may appear in the filing
but belong to a different period or entity. Because related figures
often occur in the same passage or table, the resulting answer can still appear
precise and well supported while expressing the wrong relation
\citep{mansouri2026finrank,jiang2026finrate,stolfo2024groundedness}.

A natural response is to ask the model to verify its draft against the evidence
and revise any detected error
\citep{madaan2023selfrefine,zhao2023verifyedit,dhuliawala2024cove,gou2024critic}.
However, self-verification can
introduce errors of its own
\citep{huang2024selfcorrect,kamoi2024survey}. We explore two main risks:
(1) \textit{over-editing}, in which the verifier changes correct content while
fixing a local error, and (2) \textit{misdirected verification}, in which it
checks the wrong relation or evidence and consequently preserves the original
error or adds an irrelevant claim. Reliable repair must therefore decide
\textbf{whether to repair, what to check and where to inspect, and which edit contract to apply}.

\begin{figure*}[t]
    \centering
    \includegraphics[width=\textwidth]{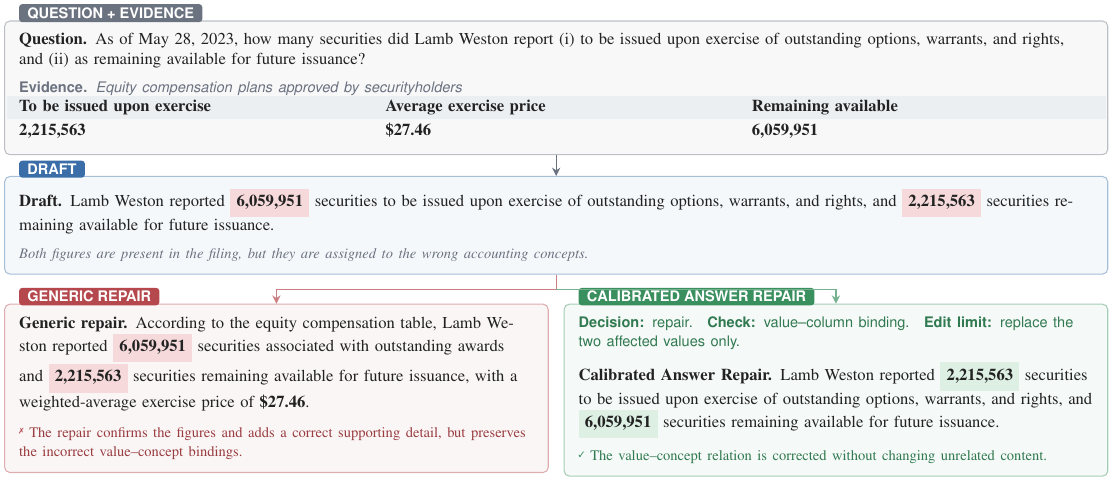}
    \caption{A value--concept binding error: both values appear in the filing,
    but the draft assigns them to the wrong accounting concepts. Generic
    verification checks presence without confirming what each value refers to.}
    \label{fig:motivating-example}
\end{figure*}

Figure~\ref{fig:motivating-example} illustrates the problem. The draft contains
both values reported in the filing, but assigns each value to the other column.
A generic verifier confirms that both numbers occur in the evidence and
therefore leaves the swap uncorrected. It may even add a valid but irrelevant
detail, supporting the current answer without fixing the error.

This example shows why confirming the presence of a number is insufficient:
the verifier must check what that number refers to. Revision carries a second
risk when it changes parts of the draft that were already correct. \method{}
addresses both risks by deciding when to repair, what relation to inspect, and
how narrowly to edit.

Motivated by this trade-off, we propose \textbf{Calibrated Answer Repair}
(\method{}). It first estimates whether a draft is likely to benefit from
revision and leaves it unchanged when repair is unlikely to help. If revision
is worthwhile, it tells the model what to check and where to look, pointing to
relevant source text or the sentence or calculation in the draft. It then
limits edits to adding supported information or correcting only a confirmed
error, leaving unrelated content unchanged.

We evaluate \method{} on FinRank, Fin-RATE, and FinQA, which cover
evidence grounding, multi-document financial analysis, and numerical reasoning. Across three
open-weight models, \method{} outperforms \draft{}, generic verification, and
\cove{} in all nine dataset--model settings, and \scverify{} and \rarr{} in
eight. At the
system level, it recovers 14.6\% of initially missed information while damaging
2.1\% of initially correct information. Its token cost is 35.2--55.3\% lower than the iterative revision
alternatives. A blinded audit further finds
lower rates of over-editing, misdirected verification, and newly introduced
contradictions. Our contributions are:
\begin{enumerate}[leftmargin=*,itemsep=2pt,topsep=2pt]
    \item We derive a break-even rule that balances the information a repair
    can recover against the correct content it may damage.
    \item We introduce \method{}, which combines a calibrated repair gate,
    targeted verification, and bounded editing.
    \item We evaluate repair across three financial QA benchmarks and test
    routing, evidence localization, and edit scope through controlled
    ablations in two dataset--model settings.
\end{enumerate}
\section{Related Work}
\label{sec:related}

\paragraph{Financial question answering.}
Financial QA benchmarks cover a range of evidence structures and reasoning demands. TAT-QA combines numerical reasoning over tables and text, FinQA provides program supervision for multi-step calculations, and ConvFinQA extends financial reasoning to conversational questions \citep{zhu2021tatqa,chen2021finqa,chen2022convfinqa}. FinRank focuses on answer grounding and provenance within released per-question candidate pools containing supporting passages and hard negatives, an emphasis shared by evidence-based QA evaluations of source quality and answer attribution~\citep{gao2023alce,schimanski2024evidenceqa}. It is not a full-filing retrieval setting. Fin-RATE evaluates detailed reasoning, cross-company comparison, and longitudinal analysis over financial filings using benchmark-provided gold context \citep{mansouri2026finrank,jiang2026finrate}. Together, these benchmarks span confusable-evidence grounding, multi-document financial analysis, and numerical reasoning.

\paragraph{Post-hoc verification and repair.}
Existing methods improve generated answers through feedback, evidence gathering, or structured verification. Self-Refine alternates model-generated feedback and revision \citep{madaan2023selfrefine}. Verify-and-Edit and \rarr{} retrieve external evidence before editing reasoning chains or unsupported claims \citep{zhao2023verifyedit,gao2023rarr}. Chain-of-Verification generates dedicated questions for checking an initial response, while CRITIC incorporates feedback from external tools \citep{dhuliawala2024cove,gou2024critic}. Sampling-based methods additionally use agreement or disagreement across generations as a reliability signal \citep{wang2023selfconsistency,manakul2023selfcheckgpt}. Recent studies also examine the conditions under which models can successfully correct their own outputs \citep{huang2024selfcorrect,kamoi2024survey}. Building on this literature, \method{} treats verification as a selective repair process that jointly determines whether to edit, what relation to check, and how much of the draft to change.

\paragraph{Selective prediction and fine-grained evaluation.}
Selective prediction allows a system to abstain when its confidence is insufficient, while model cascades allocate additional computation to selected examples \citep{geifman2017selective,kamath2020selective,varshney2022model}. These approaches motivate using additional computation only where it is likely to improve the output. \method{} applies this principle after an answer has been generated, choosing between preserving the draft and invoking repair. Fine-grained evaluation provides a complementary perspective: FActScore decomposes long-form outputs into atomic claims so that supported and unsupported content can be assessed separately \citep{min2023factscore}. We similarly assess answer requirements before and after repair, distinguishing recovered requirements from regressions that aggregate answer accuracy can obscure.
\section{When Is Repair Worthwhile?}
\label{sec:problem}

To formalize when repair is worthwhile, we score answers against atomic
\emph{answer requirements}: facts a complete answer must state correctly,
including what each value refers to. In Figure~\ref{fig:motivating-example},
both values appear but are assigned to the wrong relations, so neither
requirement is met. \emph{Requirement coverage} is the fraction of requirements
satisfied, and the missing or incorrect fraction is the \emph{miss rate} $m$.
The rubric for each question lists these requirements. Rubrics provide
supervision during development. At test time, they are used only by the
evaluator and are never shown to \method{}.

\subsection{A Break-Even Rule}

Repair can recover missed requirements but may also damage those already
satisfied. Let $f$ be the recovery rate (the fraction of missed requirements
recovered) and $d$ the damage rate (the fraction of satisfied requirements
lost). For each dataset--model setting, we estimate these rates by applying the repair to every calibration draft and comparing requirement satisfaction
before and after repair.

For miss rate $m$, the expected change in requirement coverage is
\begin{equation}
G(m)=mf-(1-m)d.
\label{eq:expected-gain}
\end{equation}
Repair is worthwhile when $G(m)\geq 0$, which gives
\begin{equation}
m \geq m^\star
=
\frac{d}{f+d}.
\label{eq:threshold}
\end{equation}
The threshold falls as recovery improves and rises as damage increases, so an error alone does not make repair worthwhile.
Because $f$ and $d$ are averages estimated on calibration data, this is a
calibrated decision rule for expected coverage on new questions, not a
guarantee of positive gain on every individual question.

\subsection{Turning the Rule into a Gate}

At test time, the rubric and miss rate are unavailable. \method{} predicts
$\widehat m$ from the question, evidence, draft, and a non-editing,
evidence-grounded probe, using rubric-based miss rates from calibration drafts
as supervision. The probe runs on every question, whether or not repair
follows.

At test time, \method{} opens the repair gate when
\begin{equation}
\widehat m \geq m^\star,
\label{eq:gate}
\end{equation}
and otherwise returns the draft unchanged. The threshold is determined by the
observed recovery and damage rates rather than tuned as an additional
hyperparameter. Because the probe has already inspected the evidence, the gate
decides whether to proceed from diagnosis to targeted repair, not whether to
perform verification. Section~\ref{sec:method} describes check selection and
the permitted edit.
\section{Calibrated Answer Repair}
\label{sec:method}

\begin{figure*}[t]
\centering
\includegraphics[width=\textwidth]{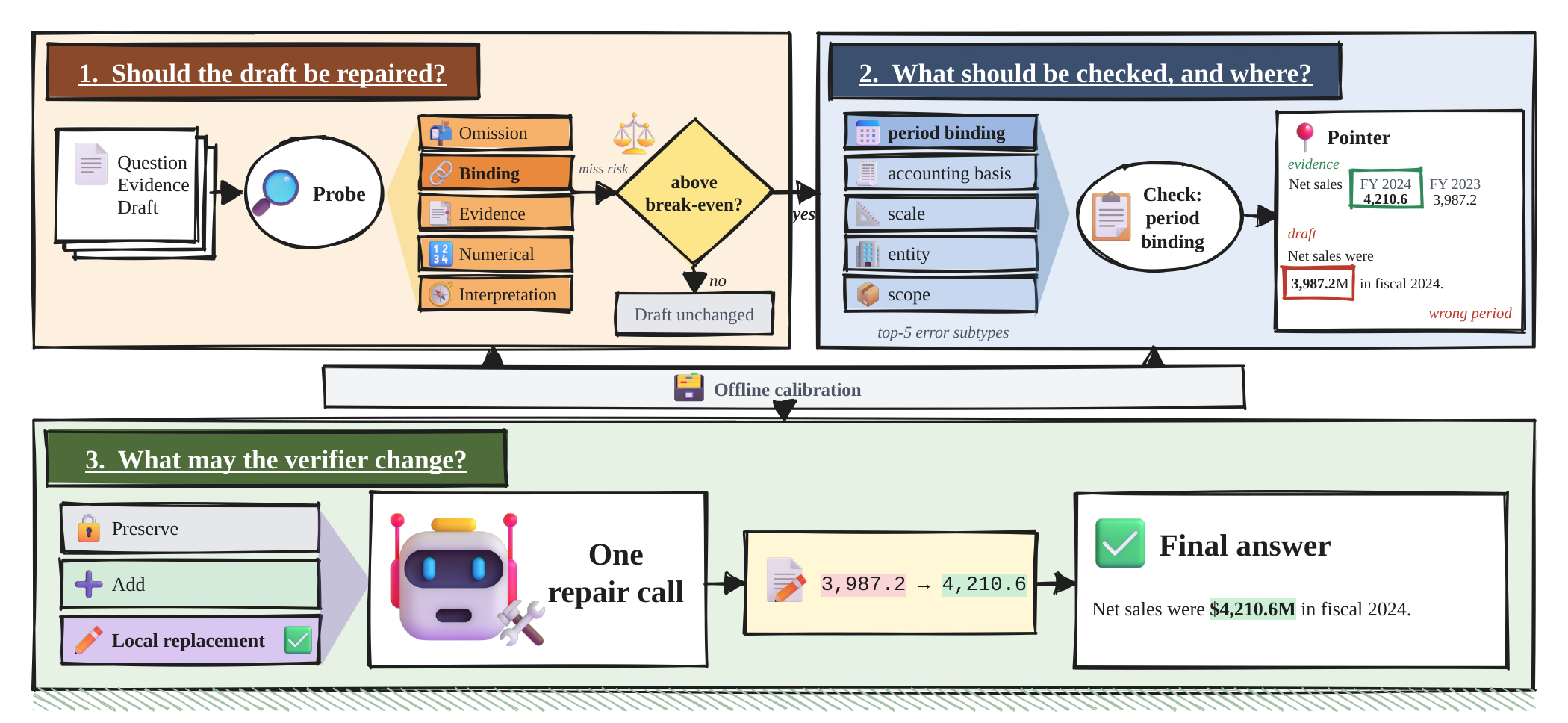}
\caption{Overview of \method{}: diagnosis gates repair, predicted error subtypes guide the check and pointer, and an edit limit constrains one repair call.}
\label{fig:overview}
\end{figure*}

\method{} structures repair around three decisions: whether it is worthwhile,
what to check and where, and how much of the answer to change. These decisions
address unnecessary revision, misdirected verification, and over-editing.
Figure~\ref{fig:overview} illustrates them with a period-binding error: the
draft assigns FY~2023 net sales to FY~2024, and a targeted local repair corrects
the value using the supplied evidence while preserving the rest of the draft.

\subsection{Learning from Calibration Errors}
\label{sec:offline}

\method{} learns from labeled calibration data. The draft model first answers
each calibration question, and the rubric identifies which answer requirements
are missing or incorrect. For each unmet requirement, a fixed language model
uses the question, evidence, draft, and requirement to write a short
description of the error.

We organize these descriptions at two levels. Five broad error families
capture the main questions a verifier may need to ask. Within each family,
similar calibration errors are grouped and named once, producing 25 more
specific error subtypes. The probe predicts the five broad families, while the
subtypes are used later to choose specific checks. The resulting error types
are fixed before test time and listed in Appendix~\ref{app:taxonomy}.

The same calibration data show how repair changes an answer. We repair every
calibration draft and compare its requirements before and after repair to
estimate the overall recovery and damage rates. Additional trials vary the
verification check and allowed edit, showing which checks recover each subtype
and how often each edit limit damages correct content. Rubric-based miss rates
provide the supervision for the diagnostic predictor. These estimates are
computed separately for each dataset--model setting.

\subsection{Should the Draft Be Repaired?}
\label{sec:gate}

Because the rubric and true miss rate are unavailable at test time, a
non-editing, evidence-grounded probe examines the question, evidence, and draft
and returns five binary judgments, one for each broad error family. A ridge
regression model combines these judgments with input features to estimate the
draft's miss rate $\widehat m$.

If $\widehat m$ is below the break-even point $m^\star$, \method{} returns the
draft unchanged. Otherwise, it proceeds from diagnosis to targeted repair
according to Equation~\ref{eq:gate}, and the gate controls whether the answer may
be edited, not whether evidence is inspected.

\subsection{What Should Be Checked, and Where?}
\label{sec:pointer}

The probe examines the current question, evidence, and draft. Its five
family-level judgments, combined with text features, estimate the likely
missed share for the repair gate, not which check to run. Separately, error
labels from semantically similar calibration questions suggest which relation
may be wrong. The selected check tests that relation, while a pointer directs
inspection to the relevant span in the current draft or supplied evidence.
Calibration-measured check recovery weights the subtype hypotheses.
The resulting weights determine check selection within a fixed budget. Retrieval, smoothing, check-scoring, and
pointer-construction details appear in Appendices~\ref{app:taxonomy},
\ref{app:policies}, and~\ref{app:pointers}.

\subsection{What May the Verifier Change?}
\label{sec:execution}

Even a well-targeted check can damage correct content if it triggers a full
rewrite. \method{} preserves the draft when no edit has positive expected
gain. Otherwise, it selects an addition or local replacement according to the
suspected error and pointer. For an omission, it inspects the source passage
and permits addition only. For binding errors, it inspects the draft span
against the source row or column. For evidence errors, it inspects the claim
and its strongest supporting passage. For numerical errors, it inspects the
calculation span and operands. For interpretation errors, it inspects the
requested relation and evidence.
These latter checks permit only local replacement.

For each action, calibration trials provide a recovery rate and a damage rate.
\method{} applies the expected-gain rule from
Equation~\ref{eq:expected-gain}, using the predicted miss rate in place of the
unknown true rate. It selects the action with the largest positive expected
gain and preserves the draft if none is expected to help.

In Figure~\ref{fig:motivating-example}, the selected check asks which value
belongs to each column, the pointer identifies the affected sentence and table
columns, and the edit is limited to replacing the two swapped values.

The selected checks, pointer, and edit action are combined in one repair call.
The prompt format and edit-validation procedure are detailed in
Appendix~\ref{app:policies}. Each question receives one diagnostic probe, and
only those passing the gate receive a repair call.
\section{Experimental Setup}
\label{sec:setup}

\subsection{Benchmarks and Primary Metrics}

Our evaluation covers three financial question-answering benchmarks with
complementary evidence and reasoning demands:
\begin{itemize}[leftmargin=*,itemsep=4pt]
    \item \textbf{FinRank}~\citep{mansouri2026finrank} evaluates answer grounding and provenance over per-question evidence. We report \textbf{Grounded Answer Accuracy (GAA)}, which counts an answer as correct only when its content is correct and at least one returned passage identifier belongs to the gold support set. This joint criterion measures both answer quality and evidence attribution, rather than passage ranking alone.
    \item \textbf{Fin-RATE}~\citep{jiang2026finrate} covers detailed reasoning (DR-QA), cross-company comparison (EC-QA), and longitudinal tracking (LT-QA). We use its gold-context setting, giving every method the same benchmark-provided supporting context. The official evaluator assigns each answer one of four labels---Correct, Partial, Incorrect, or Failure. \textbf{Correctness Rate (CR)} is the percentage labeled Correct, and we report it by subtask and use their unweighted mean as a summary.
    \item \textbf{FinQA}~\citep{chen2021finqa} tests numerical reasoning over financial text and tables. We report \textbf{Numerical Answer Accuracy (NAA)}, the proportion of normalized final answers that match the gold execution result. NAA directly evaluates the requested numerical result without requiring a generated program.
\end{itemize}

Further evidence-pool, input-cap, and normalization details appear in
Appendices~\ref{app:full-results} and~\ref{app:pointers}.

\subsection{Models and Data Partitions}

Answer-generation and repair calls use Qwen3-8B and Qwen3-14B
\citep{yang2025qwen3}, and Mistral-Small-24B-Instruct-2501
\citep{mistral2025small3}. Within each dataset--model configuration, the same
model generates the shared draft and handles subsequent diagnostic,
verification, and repair calls. Model weights remain fixed. Serving and
decoding details are provided in Appendix~\ref{app:policies}. Table~\ref{tab:splits}
summarizes the calibration, development, and held-out test partitions.

\begin{table}[t]
\centering
\footnotesize
\begin{tabular}{@{}lccc@{}}
\toprule
Partition & FinRank & Fin-RATE & FinQA\\
\midrule
Calibration & 910 & 5,250 & 6,251\\
Development & 140 & 750 & 883\\
Held-out test & 135 & 1,500 & 1,147\\
\bottomrule
\end{tabular}
\caption{Partition sizes by benchmark.}
\label{tab:splits}
\end{table}

FinRank uses the released held-out-ticker split. Fin-RATE uses fixed 70/10/20
splits within each subtask, grouped by filing bundle (1,750/250/500 questions
per subtask). FinQA uses its official split.

Fin-RATE uses the official three-judge fusion of GPT-5, DeepSeek-V3.2, and
Qwen3-235B. Labels map to 0 (Failure), 1 (Incorrect), 3 (Partial), and 5
(Correct) before fusion. Weights, ordered as DeepSeek/Qwen/GPT-5, are
0.70/0.10/0.20 for DR-QA and 0.14/0.12/0.74 for EC-QA and LT-QA. Only a fused
Correct label counts toward CR.
This benchmark evaluator is separate from the requirement judge. We use
Claude Sonnet 5~\citep{anthropic2026sonnet5} to label each
answer-requirement/candidate-answer pair with the prompt in
Appendix~\ref{app:policies}. Those labels measure requirement transitions
and support calibration of the gate and edit actions.
On 400 stratified requirement decisions, Claude's labels agree with adjudicated human judgments in 91.8\% of cases ($\kappa=0.83$).
Calibration data
provide labels and paired repair outcomes, development data select non-test
hyperparameters, and held-out data provide benchmark scores. Gold answers,
requirements, and scoring annotations are reserved for evaluation or offline
calibration. The Fin-RATE gold context is the shared inference-time evidence.

\subsection{Comparison Methods}

\begin{itemize}[leftmargin=*,itemsep=2pt]
    \item \draft{} returns the shared initial answer without post-draft processing.
    \item \verify{} applies one generic evidence-grounded check-and-revise pass.
    \item \scverify{} generates three independent reviews and returns the majority-voted normalized answer. Aggregation and tie-breaking are specified in Appendix~\ref{app:policies}.
    \item \cove{}~\citep{dhuliawala2024cove} uses factored verification: it answers draft-generated questions independently from evidence alone, then revises the shared draft.
    \item \rarr{}~\citep{gao2023rarr} checks draft-derived claims against the shared evidence pool, localizes mismatches, and minimally edits them under its edit-distance limit.
    \item \method{} uses the calibrated gate to decide whether to proceed from diagnosis to repair, then selects a check and pointer and constrains the edit.
\end{itemize}

Implementation details for the supplied-evidence \cove{} and \rarr{} adaptations appear in Appendix~\ref{app:baseline-adaptations}.

\subsection{Evaluation and Cost}

Within each dataset--model configuration, all methods receive the same
question, evidence, and initial draft. For Fin-RATE this evidence is the
benchmark-provided gold context. Requirement coverage,
recovery, and damage follow the definitions in Section~\ref{sec:problem}.

Inference cost is the additional post-draft input and output tokens. For
\method{},
\begin{equation}
C_{\method}=C_{\mathrm{probe}}
+\mathbb{1}[\text{repair gate opens}]C_{\mathrm{repair}}.
\label{eq:cost}
\end{equation}
The evidence-grounded probe runs on every question, so its cost is included
whether or not the gate opens. The cost of \scverify{} includes all three
reviews.
Recovery, damage, calls, and token totals are macro-averaged over the
dataset--model configurations. A blinded two-annotator audit compares
\verify{} and \method{} on 144 paired drafts for over-editing, misdirected
verification, and newly introduced contradictions.
\section{Results}
\label{sec:results}

\subsection{End-to-End Performance}
\label{sec:mainresults}

Table~\ref{tab:main-results} shows that \method{} improves over \draft{}, \verify{}, and \cove{} in all nine settings, and over \scverify{} and \rarr{} in eight. It trails the latter two only on Qwen3-14B FinRank (35.56\% vs. 36.30\%).

\begin{table*}[t]
\centering
\scriptsize
\setlength{\tabcolsep}{3pt}
\renewcommand{\arraystretch}{1.1}
\begin{tabular*}{\textwidth}{@{\extracolsep{\fill}}ll*{6}{S[table-format=2.2]}@{}}
\toprule
Dataset / Metric & Model & \multicolumn{1}{c}{\draft{}} & \multicolumn{1}{c}{\verify{}} & \multicolumn{1}{c}{\scverify{}} & \multicolumn{1}{c}{\cove{}} & \multicolumn{1}{c}{\rarr{}} & \multicolumn{1}{c}{\method{}}\\
\midrule
FinRank / GAA & Qwen3-8B & 23.70 & 22.96 & 25.19 & 25.93 & 28.15 & \multicolumn{1}{c}{\textbf{29.63}}\\
& Qwen3-14B & 31.11 & 31.85 & \multicolumn{1}{c}{\textbf{36.30}} & 34.07 & \multicolumn{1}{c}{\textbf{36.30}} & 35.56\\
& Mistral-24B & 29.63 & 30.37 & 31.11 & 31.11 & 33.33 & \multicolumn{1}{c}{\textbf{34.81}}\\
\midrule
Fin-RATE / Macro CR & Qwen3-8B & 5.53 & 5.73 & 6.13 & 6.67 & 7.47 & \multicolumn{1}{c}{\textbf{8.20}}\\
& Qwen3-14B & 11.27 & 11.60 & 12.33 & 13.07 & 14.07 & \multicolumn{1}{c}{\textbf{14.67}}\\
& Mistral-24B & 16.73 & 17.13 & 17.87 & 18.07 & 19.40 & \multicolumn{1}{c}{\textbf{20.07}}\\
\midrule
FinQA / NAA & Qwen3-8B & 63.21 & 62.77 & 64.08 & 62.95 & 65.65 & \multicolumn{1}{c}{\textbf{66.43}}\\
& Qwen3-14B & 68.35 & 68.00 & 69.22 & 70.27 & 71.32 & \multicolumn{1}{c}{\textbf{72.01}}\\
& Mistral-24B & 70.18 & 69.66 & 70.88 & 71.05 & 72.28 & \multicolumn{1}{c}{\textbf{72.97}}\\
\bottomrule
\end{tabular*}
\caption{Held-out benchmark scores (percent, with higher values indicating better performance). Bold marks the best method per setting.}
\label{tab:main-results}
\end{table*}

Fin-RATE Macro CR is the unweighted mean of the official DR-QA, EC-QA, and
LT-QA Correctness Rates. The full subtask results appear in
Appendix~\ref{app:finrate-results}.

Effects vary by task and model. On FinRank, \method{} exceeds \scverify{} by 4.44 points for Qwen3-8B and 3.70 for Mistral-24B, but trails by 0.74 for Qwen3-14B. Fin-RATE task-level gains appear in Section~\ref{sec:finrate-task-gains}. On FinQA, generic verification lowers accuracy, while \method{} gains 2.09--2.79 points over \scverify{} by avoiding changes to correct calculations.

\subsection{How Does Generic Verification Fail?}
\label{sec:verification-failures}

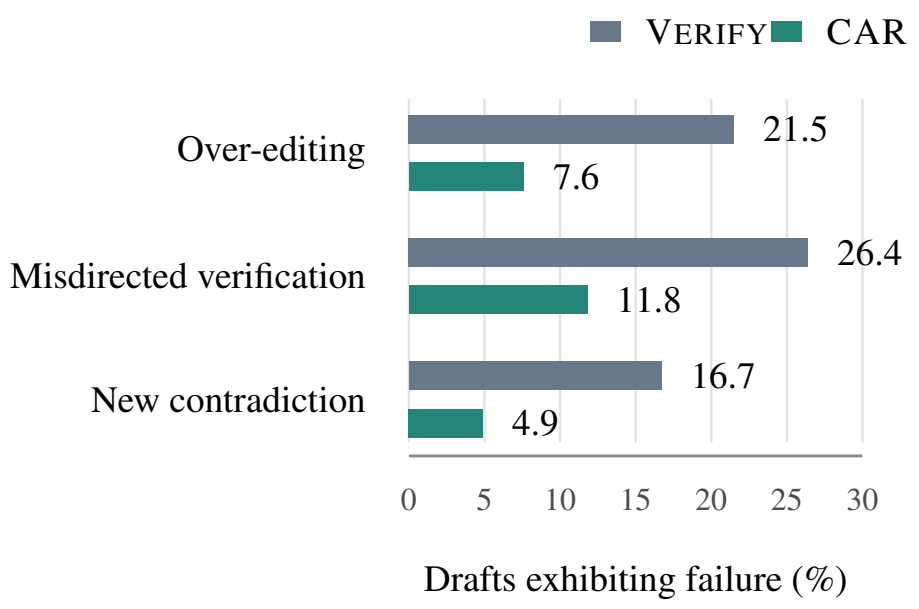
\begin{figure}[b]
\centering
\begin{tikzpicture}[x=0.1cm,y=0.68cm,font=\scriptsize]
\definecolor{verifybar}{RGB}{102,120,138}
\definecolor{carbar}{RGB}{37,133,121}

\foreach \tick in {0,5,...,30} {
  \pgfmathsetmacro{\xcoord}{40+\tick}
  \draw[gray!22,line width=0.45pt] (\xcoord,-0.82) -- (\xcoord,2.52);
  \node[anchor=north,text=black!70,font=\tiny] at (\xcoord,-1.08) {\tick};
}
\draw[black!45,line width=0.55pt] (40,-0.95) -- (70,-0.95);

\node[anchor=east] at (38.4,2.0) {Over-editing};
\node[anchor=east] at (38.4,0.8) {Misdirected verification};
\node[anchor=east] at (38.4,-0.4) {New contradiction};

\fill[verifybar] (40,2.09) rectangle (61.5,2.37);
\fill[carbar] (40,1.63) rectangle (47.6,1.91);
\fill[verifybar] (40,0.89) rectangle (66.4,1.17);
\fill[carbar] (40,0.43) rectangle (51.8,0.71);
\fill[verifybar] (40,-0.31) rectangle (56.7,-0.03);
\fill[carbar] (40,-0.77) rectangle (44.9,-0.49);

\node[anchor=west] at (62.2,2.23) {21.5};
\node[anchor=west] at (48.3,1.77) {7.6};
\node[anchor=west] at (67.1,1.03) {26.4};
\node[anchor=west] at (52.5,0.57) {11.8};
\node[anchor=west] at (57.4,-0.17) {16.7};
\node[anchor=west] at (45.6,-0.63) {4.9};

\fill[verifybar] (52,3.08) rectangle (54,3.28);
\node[anchor=west] at (54.35,3.18) {\verify{}};
\fill[carbar] (64,3.08) rectangle (66,3.28);
\node[anchor=west] at (66.35,3.18) {\method{}};
\node[anchor=center] at (55,-2.18) {Drafts exhibiting failure (\%)};
\end{tikzpicture}
\caption{Failure rates in the blinded 144 paired drafts.}
\label{fig:verification-failures}
\end{figure}

Figure~\ref{fig:verification-failures} shows that \method{} reduces all three audited failure modes, with the largest difference in misdirected verification. Inter-annotator agreement ranged from 87.8\% to 95.1\% across labels (Cohen's $\kappa=0.75$--$0.82$).

\subsection{Repair Safety and Cost}
\label{sec:safety-cost}

Table~\ref{tab:safety-cost} separates repair behavior from benchmark score. Generic \verify{} recovers 8.1\% of initially missed requirements but damages 6.7\% of those already satisfied. \scverify{} recovers 9.8\% and damages 5.5\% at three calls per question. \cove{} recovers 11.7\% and damages 4.8\% at 4.00 calls per question, while \rarr{} recovers 13.8\% and damages 2.4\% at 3.58 calls per question. At the system level, \method{} has the highest recovery rate at 14.6\% and the lowest damage rate at 2.1\%, with 1.54 calls per question. Because this aggregate includes drafts for which the gate stays closed, it does not by itself isolate the safety of the edit policy.

\begin{table*}[t]
\centering
\scriptsize
\renewcommand{\arraystretch}{1.12}
\begin{tabular*}{\textwidth}{@{\extracolsep{\fill}}l S[table-format=2.1] S[table-format=2.1] S[table-format=1.2] S[table-format=5.0] S[table-format=3.1]@{}}
\toprule
Method & \multicolumn{1}{c}{Recovery (\%)} & \multicolumn{1}{c}{Damage (\%)} & \multicolumn{1}{c}{Calls/q} & \multicolumn{1}{c}{\makecell[c]{Tokens\\per question}} & \multicolumn{1}{c}{Rel. cost (\%)}\\
\midrule
\draft{} & \multicolumn{1}{c}{\textemdash} & \multicolumn{1}{c}{\textemdash} & 0.00 & 0 & 0.0\\
\verify{} & 8.1 & 6.7 & 1.00 & 3417 & 100.0\\
\scverify{} & 9.8 & 5.5 & 3.00 & 10467 & 306.3\\
\cove{} & 11.7 & 4.8 & 4.00 & 11960 & 350.0\\
\rarr{} & 13.8 & 2.4 & 3.58 & 8240 & 241.1\\
\method{} & \multicolumn{1}{c}{\textbf{14.6}} & \multicolumn{1}{c}{\textbf{2.1}} & 1.54 & 5343 & 156.4\\
\bottomrule
\end{tabular*}
\caption{Requirement transitions and post-draft cost, macro-averaged across settings. Relative cost is normalized to \verify{} at 100.0\%.}
\label{tab:safety-cost}
\end{table*}

We also separate system-level transitions from outcomes conditional on
intervention and actual edits. Gate-open recovery and damage are measured only
on drafts sent to repair. Changed-answer damage is conditioned on an answer
change, while edit rate is the share of all questions whose final answer
differs from the draft (Table~\ref{tab:conditional-safety}).

\begin{table}[tbp]
\centering
\scriptsize
\begin{tabular*}{\columnwidth}{@{\extracolsep{\fill}}l S[table-format=2.1]@{}}
\toprule
\method{} metric & \multicolumn{1}{c}{Rate (\%)}\\
\midrule
Overall recovery & 14.6\\
Overall damage & 2.1\\
Gate-open recovery & 20.3\\
Gate-open damage & 5.3\\
Changed-answer damage & 8.4\\
Edit rate & 35.0\\
\bottomrule
\end{tabular*}
\caption{System-level and conditional repair outcomes.}
\label{tab:conditional-safety}
\end{table}

Among gate-open drafts, recovery is 20.3\% and damage is 5.3\%, while only 35.0\%
of final answers change, and 8.4\% of changed answers damage a satisfied
requirement. The lower overall damage therefore reflects both selective
intervention and outcomes after editing. These aggregate rates do not isolate
the causal contribution of the gate from that of the edit limit.

The saving is not obtained by ignoring diagnosis. Every \method{} example pays for the probe, and 54.1\% additionally invoke repair, for an average of 1.54 calls per question. Including both calls, \method{} uses 5,343 additional post-draft tokens per question. This is 156.4\% of the 3,417 tokens used by \verify{}, or 56.4\% more. It uses 49.0\% fewer tokens than \scverify{}, 55.3\% fewer than \cove{}, and 35.2\% fewer than \rarr{}.

\subsection{Component Ablations}
\label{sec:component-ablation}

Table~\ref{tab:component-ablation} reports Qwen3-14B FinRank performance when changing the gate, pointer, check selection, or edit constraint.

\begin{table}[t]
\centering
\scriptsize
\setlength{\tabcolsep}{5pt}
\renewcommand{\arraystretch}{1.1}
\begin{tabular}{@{}l S[table-format=2.2]@{}}
\toprule
Variant & \multicolumn{1}{c}{GAA}\\
\midrule
Full \method{} & 35.56\\
Always repair & 30.37\\
No pointer & 28.89\\
Fixed checklist & 31.85\\
Unbounded rewrite & 27.41\\
\bottomrule
\end{tabular}
\caption{FinRank GAA for Qwen3-14B \method{} variants.}
\label{tab:component-ablation}
\end{table}

We repeat the four single-factor ablations on FinQA with Mistral-24B to test the component effects on another dataset, model family, and a larger numerical-answer test set. The results and one-factor controls appear in Appendix~\ref{app:finqa-ablation}.
\section{Analysis}
\label{sec:analysis}

\subsection{Does Selective Repair Help?}
\label{sec:analysis-gate}

\begin{center}
\centering
\scriptsize
\setlength{\tabcolsep}{3.5pt}
\renewcommand{\arraystretch}{1.1}
\begin{tabular}{@{}l S[table-format=1.3] S[table-format=1.3] S[table-format=1.3] S[table-format=2.1]@{}}
\toprule
Dataset & \multicolumn{1}{c}{\makecell{Always-repair\\$f$}} & \multicolumn{1}{c}{\makecell{Always-repair\\$d$}} & \multicolumn{1}{c}{$m^\star$} & \multicolumn{1}{c}{\makecell{Gate-open rate (\%)}} \\
\midrule
FinRank & 0.142 & 0.095 & 0.401 & 62.7 \\
Fin-RATE & 0.115 & 0.089 & 0.436 & 55.4 \\
FinQA & 0.068 & 0.081 & 0.543 & 44.1 \\
\bottomrule
\end{tabular}
\captionof{table}{Break-even quantities averaged across models.}
\label{tab:gate-calibration}
\end{center}

The calibration pattern follows the break-even rule. On FinRank, recovery exceeds damage ($f=0.142$, $d=0.095$), giving the lowest threshold ($m^\star=0.401$) and the highest gate-open rate (62.7\%). On FinQA, damage exceeds recovery ($f=0.068$, $d=0.081$), so the threshold rises to 0.543 and the gate opens for 44.1\% of drafts. Fin-RATE falls between these cases. Relative to always applying the repair pipeline, selective repair raises requirement coverage by 1.48, 0.80, and 0.81 points on FinRank, Fin-RATE, and FinQA, respectively. Across models, thresholds range from 0.395 to 0.408 on FinRank, 0.427 to 0.453 on Fin-RATE, and 0.527 to 0.566 on FinQA. Their gate-open rates range from 59.8 to 66.7\%, 52.6 to 58.9\%, and 40.2 to 47.8\%, respectively (Appendix~\ref{app:calibration}). The separated ranges show greater threshold variation across benchmarks than across models within a benchmark.

We compare miss-rate predictors using text features, probe verdicts, and their combination (Table~\ref{tab:gate-predictor}).

\begin{center}
\centering
\scriptsize
\renewcommand{\arraystretch}{1.1}
\begin{tabular}{@{}l S[table-format=1.3] S[table-format=1.3]@{}}
\toprule
Miss-rate input & \multicolumn{1}{c}{MAE $\downarrow$} & \multicolumn{1}{c}{Pearson $r$ $\uparrow$}\\
\midrule
Dataset mean only & 0.196 & \multicolumn{1}{c}{\textemdash}\\
Text features only & 0.171 & 0.344\\
Probe verdicts only & 0.158 & 0.473\\
Text + probe & \textbf{0.146} & \textbf{0.512}\\
\bottomrule
\end{tabular}
\captionof{table}{Miss-rate prediction on grouped development folds.}
\label{tab:gate-predictor}
\end{center}

Adding the five evidence-grounded verdicts reduces MAE from 0.171 to 0.146, a
14.6\% reduction, and raises Pearson correlation from 0.344 to 0.512 versus
text features alone. Probe verdicts alone also outperform text features, while
combining both gives the lowest error. These results support using the probe
to diagnose each draft, separately from the later gate decision to permit an
edit.

\subsection{Which Repair Components Matter?}
\label{sec:targeted-analysis}

On Qwen3-14B FinRank, question-specific routing adds 3.71 GAA points over the fixed checklist (35.56 vs. 31.85). Always repair, no pointer, and unbounded rewrite score 30.37, 28.89, and 27.41, respectively, making localization and edit scope the larger contributors in this setting. The FinQA--Mistral-24B ablation shows the same ordering on numerical answer accuracy: removing the pointer costs 2.70 points and removing the edit bound costs 4.01, compared with 2.44 for always repairing. The fixed-checklist gap is 0.87 points, small relative to the pointer and edit-scope effects (Appendix~\ref{app:finqa-ablation}).

\subsection{Do Pointers Localize Relevant Content?}

Among omissions, source pointers place a supporting passage in the top three
for 72.8\% of FinRank cases and 52.4\% of Fin-RATE cases, compared with 46.9\%
and 20.1\% under random ranking. On FinQA numerical failures, calculation
pointers overlap the reference span in 68.9\% of cases and localize all required
operands in 51.7\%, versus 41.6\% and 27.4\% with answer-sentence pointers.
For Fin-RATE, penalizing passages that resemble the draft lowers source Hit@3
from 52.4\% to 29.2\%. The drop suggests that an omitted fact often appears in
context topically related to content already in the draft. Question relevance
is therefore a better ranking signal here than draft novelty. Full definitions
and breakdowns appear in Appendix~\ref{app:pointers}.

\subsection{How Do Fin-RATE Gains Vary by Task?}
\label{sec:finrate-task-gains}

\method{} improves all Fin-RATE model--subtask pairs. Absolute gains are
largest on DR, while EC and LT also improve from lower baselines. Table~\ref{tab:finrate-gains}
reports model-level point changes: DR gains range from 4.4 to 5.4, EC from
1.4 to 2.8, and LT from 2.2 to 2.8. For Qwen3-8B, EC rises from 0.4 to 1.8.
Point changes are more informative than relative percentages from a near-zero
starting value. Full Correctness Rates appear in Appendix~\ref{app:finrate-results}.

\begin{center}
\centering
\scriptsize
\setlength{\tabcolsep}{5pt}
\renewcommand{\arraystretch}{1.1}
\begin{tabular}{@{}l S[table-format=1.1] S[table-format=1.1] S[table-format=1.1]@{}}
\toprule
Model & \multicolumn{1}{c}{DR-QA} & \multicolumn{1}{c}{EC-QA} & \multicolumn{1}{c}{LT-QA}\\
\midrule
Qwen3-8B & +4.4 & +1.4 & +2.2\\
Qwen3-14B & +5.4 & +2.0 & +2.8\\
Mistral-24B & +4.8 & +2.8 & +2.4\\
\bottomrule
\end{tabular}
\captionof{table}{Fin-RATE \method{} minus \draft{} CR.}
\label{tab:finrate-gains}
\end{center}
\section{Conclusion}

\method{} makes repair selective, targeted, and bounded. It outperforms standard verification approaches in all nine benchmark--model settings and stronger verification variants in eight. Its post-draft token cost is higher than a single verification pass but lower than the iterative revision alternatives. Controlled ablations show that evidence localization and bounded edits matter across both judge-based and numerical-answer evaluation, while the fixed-checklist gap is small on FinQA.

\clearpage
\section*{Limitations}

Our evaluation covers three English-language financial benchmarks and three open-weight models. Results may not transfer directly to other domains, languages, or model families, and the calibrated repair policy relies on paired development outcomes. Fin-RATE uses benchmark-provided gold context, so our experiments focus on answer repair rather than end-to-end retrieval. Requirement-level outcomes use an independent judge, while benchmark metrics and a blinded audit provide complementary checks.

\section*{Ethical Considerations}

Higher requirement coverage does not make a financial answer safe for investment, accounting, or regulatory use. Deployed systems should expose their evidence, avoid unsupported repairs, and retain qualified human review for consequential decisions.

The benchmarks are based on public financial disclosures, but redistribution remains subject to the corresponding dataset licenses. Failure annotation, calibration, and repeated model inference also incur compute and data-transfer costs.

\paragraph{Writing assistance.}
A generative AI assistant helped organize and edit the manuscript and check internal consistency.

\clearpage
\bibliographystyle{icml2025}
\bibliography{references}

\clearpage
\appendix
\section{Failure Families and Induction}
\label{app:taxonomy}

\paragraph{Annotation source.}
For each missed calibration requirement, a frozen LLM annotator receives the question, supplied evidence, draft, and missed requirement and writes a short failure description that abstracts away company names, products, dates, and values. The annotator is used only on calibration data. We manually review a subset to ensure that the descriptions identify the failed relation without copying instance-specific content.

The annotation instruction is:
\promptbox{%
Describe why the draft fails to state the missed answer requirement. Name the
missing or incorrect relation, but abstract away instance-specific entities,
dates, products, and values. Do not propose a repair and do not infer failures
beyond the supplied requirement.%
}

\paragraph{Fixed families, induced leaves.}
The hierarchy contains five fixed families and 25 fine-grained leaves induced from the pooled calibration descriptions. We embed the descriptions, cluster them bottom-up, assign canonical names, and merge clusters that receive the same name. The resulting hierarchy is frozen before test evaluation and shared across dataset--model settings. Table~\ref{tab:failure-families} summarizes the five families. Table~\ref{tab:failure-leaves} gives all leaves used for estimation and routing.

\begin{table*}[t]
\centering
\small
\begin{tabularx}{\textwidth}{@{}lXX@{}}
\toprule
Family & Definition & Typical checks and localization\\
\midrule
Omission & A requested entity, period, qualifier, comparison, or result is absent. & Retrieve the supporting source passage and append only the supported missing relation.\\
Binding & A stated value is attached to the wrong entity, period, unit, or accounting concept. & Trace the value to its row, column, and passage, then locally replace the incorrect binding.\\
Evidence & A claim is unsupported, weakly supported, or grounded in a distractor passage. & Compare the draft sentence with its strongest evidence and verify provenance.\\
Numerical & The selected operands, operation, scale, sign, or computed result is wrong. & Point to the calculation span and recompute from cited inputs.\\
Interpretation & The answer misstates a comparison, trend, aggregation, or requested relation. & Re-read the question relation and check the conclusion against the evidence.\\
\bottomrule
\end{tabularx}
\caption{The five fixed failure families. Fine-grained leaves are induced from calibration data and used for hierarchical smoothing and routing.}
\label{tab:failure-families}
\end{table*}

\begin{table*}[t]
\centering
\scriptsize
\begin{tabularx}{\textwidth}{@{}llX@{}}
\toprule
Family & Leaf & Definition\\
\midrule
Omission & \texttt{f\_omit\_narrative} & A requested fact appears in narrative evidence but is absent from the answer.\\
& \texttt{f\_omit\_absent} & The requested fact is not supported by the supplied evidence and cannot be recovered by verification.\\
& \texttt{f\_omit\_derivable} & The required inputs are present, but the answer omits a value that must be derived.\\
& \texttt{f\_omit\_table} & A requested value appears in a supplied table but is not reported.\\
& \texttt{f\_omit\_clause} & A multi-part question is answered only in part.\\
\midrule
Binding & \texttt{f\_entity} & A value is attributed to the wrong company or entity.\\
& \texttt{f\_concept} & A related but different accounting concept is substituted for the requested one.\\
& \texttt{f\_period} & A value from the wrong fiscal year or quarter is used.\\
& \texttt{f\_unit} & The currency, magnitude, or percentage-point unit is wrong.\\
& \texttt{f\_scope} & Consolidated, segment, or regional reporting scopes are confused.\\
& \texttt{f\_version} & A superseded value is used instead of its restated version.\\
& \texttt{f\_basis} & GAAP and adjusted, reported and constant-currency, or similar bases are mixed.\\
\midrule
Evidence & \texttt{f\_unsupported} & The answer asserts a claim not supported by the supplied evidence.\\
& \texttt{f\_wrong\_evidence} & A value or claim is taken from the wrong passage or table.\\
& \texttt{f\_cross\_source} & Conflicting sources are combined without resolving the inconsistency.\\
\midrule
Numerical & \texttt{f\_scale} & Thousands, millions, percentages, or percentage points are confused.\\
& \texttt{f\_arithmetic} & The intended operation is correct but its arithmetic result is wrong.\\
& \texttt{f\_operation} & The wrong operation is used, such as a difference instead of percentage change.\\
& \texttt{f\_sign} & The direction of change or numerical sign is reversed.\\
& \texttt{f\_aggregation} & Components are double-counted or a subtotal is added to its own parts.\\
& \texttt{f\_denominator} & A ratio or growth rate uses the wrong denominator.\\
\midrule
Interpretation & \texttt{f\_unsupported\_explanation} & The stated explanation or reason is not given by the evidence.\\
& \texttt{f\_definition} & A financial term is defined or applied incorrectly.\\
& \texttt{f\_wrong\_driver} & A secondary driver is emphasized while the dominant driver is omitted.\\
& \texttt{f\_causal\_overclaim} & Correlation or co-occurrence is overstated as a causal relation.\\
\bottomrule
\end{tabularx}
\caption{The complete 25-leaf taxonomy. Frequencies are computed from the frozen calibration-label export rather than from test outputs. The pooled calibration corpus contains 5,019 labeled errors.}
\label{tab:failure-leaves}
\end{table*}

In the frozen calibration export, Evidence leaves account for 5.4\% of labeled errors, and \texttt{f\_causal\_overclaim} is the only zero-observation leaf. The accompanying label manifest reports the exact count and percentage for every leaf. Test labels are not used to estimate these frequencies.

\paragraph{Annotation record.}
Each calibration record contains the question identifier, answer requirement, draft, satisfaction and recoverability judgments, generated failure description, induced leaf and family, reference error span when existing content is wrong, and supporting-passage set. We retain satisfied and unrecoverable requirements so that miss rates and population-level transitions are computed over the complete cohort.

\paragraph{Question-conditioned leaf distribution.}
At test time, a fixed \texttt{all-MiniLM-L6-v2} encoder embeds the first 700 characters of the new question as a normalized 384-dimensional vector. We compare it by cosine similarity with the calibration questions and retrieve the 50 nearest questions. The retrieval unit is the question: all failure labels attached to each neighbor contribute to the local counts, so a question with multiple missed requirements contributes multiple labels. No LLM is used in this retrieval stage.

Let $n_g(q)$ and $n_\ell(q)$ be the neighborhood counts for family $g$ and leaf $\ell$, and let $p_g$ and $p_{\ell\mid g}$ be their global calibration frequencies. We smooth at both levels,
\begin{align}
P(g\mid q)&=\frac{n_g(q)+\tau_g p_g}{\sum_{g'}n_{g'}(q)+\tau_g},\\
P(\ell\mid g,q)&=\frac{n_\ell(q)+\tau_\ell p_{\ell\mid g}}{n_g(q)+\tau_\ell},
\end{align}
and set $\pi_\ell(q)=P(g(\ell)\mid q)P(\ell\mid g(\ell),q)$. We use $\tau_g=2$ and $\tau_\ell=4$ and pass the five highest-mass leaves to the repair planner. The neighborhood size was selected on the development split from $\{20,50,120\}$ using top-$k$ leaf recall, and the choice was stable across this range. Question features used by the gate are deliberately excluded from semantic retrieval.

\paragraph{Pooling and stability.}
Leaf-specific check efficacy is shrunk toward its parent-family estimate. We induce the hierarchy once from pooled calibration descriptions so that leaf identities remain comparable across the nine settings. The hierarchy, sentence encoder, neighbor count, and smoothing strengths are frozen before test evaluation.
\section{Diagnostic Probe}
\label{app:probe}

The probe receives the question, the same supplied evidence blocks and
identifiers shown to the draft, the draft, and the fixed family definitions.
Its instruction and output format are:

\promptbox{%
Inspect the draft without rewriting it. For each of the five error families,
decide whether the question and supplied evidence provide a concrete reason to
suspect that the draft omits or misstates information requested by the
question. Return one verdict and a short evidence-grounded rationale per
family. A verdict must be exactly \texttt{yes} or \texttt{no}. Do not propose
edits, use outside information, or assume access to an evaluation rubric.\\
\textbf{Question:} \texttt{<question>}\\
\textbf{Evidence:} \texttt{<same evidence blocks and identifiers as Draft>}\\
\textbf{Draft:} \texttt{<draft>}\\
\textbf{Error-family definitions:} \texttt{<five fixed definitions>}\\
\textbf{Output format:} Emit exactly five records in this fixed order:
\texttt{Omission: <yes|no>, rationale: <reason>},
\texttt{Binding: <yes|no>, rationale: <reason>},
\texttt{Evidence: <yes|no>, rationale: <reason>},
\texttt{Numerical: <yes|no>, rationale: <reason>},
\texttt{Interpretation: <yes|no>, rationale: <reason>}.%
}

The predictor converts \texttt{yes} to 1 and \texttt{no} to 0, in the fixed
order Omission, Binding, Evidence, Numerical, Interpretation. The miss-rate
model combines this five-value vector with textual features of $(q,x,y^0)$:
question length and number of requested relations, named entities and
periods, evidence length and structure, draft length, and whether the draft
explicitly states relevant entities, dates, quantities, and comparison terms.

We select the ridge regularization strength with five question-grouped folds, keeping all examples derived from the same question together. The target is the calibration miss rate computed from answer requirements. We search $\alpha\in\{10^{-4},10^{-3},10^{-2},10^{-1},1,10,100\}$ on the development split. Neither the target nor the rubric is available to the model at test time. The gate threshold is not tuned separately. It is fixed by Equation~\ref{eq:threshold}.
\section{Pointer Construction}
\label{app:pointers}

The pointer candidate universe matches the evidence available to every end-to-end method. For FinRank, candidates are the released per-question supporting passages and hard negatives. For Fin-RATE, we segment the benchmark-provided gold context at document and table boundaries into overlapping, length-bounded chunks. Pointers rank only within this shared context and perform no additional retrieval. For FinQA, candidates come from the supplied \texttt{pre\_text}, linearized table, and \texttt{post\_text}. The \texttt{gold\_inds} are retained only for evaluation. A fixed off-the-shelf text encoder embeds the question and candidates, and cosine similarity ranks candidates by question relevance. The encoder is not fine-tuned on any evaluation benchmark. We fix the encoder, chunking, and $k$ on the development split.

For omissions, the repair call receives the top-$k$ passages ranked by question relevance. We also evaluate the novelty-penalized retrieval score
\begin{equation}
s_{\mathrm{novel}}(c)
=\operatorname{sim}(c,q)
-\lambda\max_{u\in y^0}\operatorname{sim}(c,u).
\end{equation}
as a pointer-ranking ablation.

For wrong content, a frozen off-the-shelf support scorer evaluates every draft-sentence and passage pair. It is not trained on FinRank, Fin-RATE, or FinQA and receives no gold support annotations at test time. For each sentence, the scorer selects its best-matching passage and ranks the pair by how poorly that passage supports the sentence. Reference error spans are the minimal draft spans marked during calibration annotation as asserting the incorrect value, entity, period, or concept. We report sentence-overlap accuracy and joint sentence--passage accuracy. Encoder and scorer configurations are fixed before test evaluation, and their local compute is reported separately from generative token cost.

For FinQA, the test-time pointer is constructed from the question, evidence, and numerical mentions in the natural-language draft. Gold program operands and supporting facts are used only for evaluation: after alignment to draft mentions, the smallest expression or sentence containing the answer and required operands defines the reference calculation span. A pointer is correct when its marked expression overlaps this span. Operand localization additionally requires all operands used by the gold program.

\begin{table}[t]
\centering
\scriptsize
\setlength{\tabcolsep}{3pt}
\begin{tabular}{@{}llcc@{}}
\toprule
Setting & Metric & Pointer & Comparison\\
\midrule
FinRank & Source Hit@3 & 72.8 & 46.9 random\\
Fin-RATE & Source Hit@3 & 52.4 & 20.1 random\\
Fin-RATE & Source Hit@3 & 52.4 relevance & 29.2 novelty\\
FinQA & Span overlap & 68.9 & 41.6 answer sent.\\
FinQA & All operands & 51.7 & 27.4 answer sent.\\
\bottomrule
\end{tabular}
\caption{Pointer localization results. Values are percentages. ``Novelty'' subtracts similarity to the draft from question relevance.}
\label{tab:pointer-results}
\end{table}
\section{Hierarchical Calibration}
\label{app:calibration}

For a candidate check $a$ and edit policy $\eta$, let $y^{a,\eta}$ be its repaired calibration answer, $M_\ell(y^0)$ the missed requirements assigned to leaf $\ell$, and $C(y^0)$ the requirements already satisfied. Single-check recovery and policy-specific damage are
\begin{align}
e_{a\ell\eta}
&=P(J(z,y^{a,\eta})=1\mid z\in M_\ell(y^0)),\\
d_{a,\eta,b}
&=P(J(z,y^{a,\eta})=0\mid z\in C(y^0),\eta,b),
\end{align}
where $\eta$ is the edit policy and $b$ is the editable fraction of the draft.

We stabilize sparse check--leaf pairs with a regularized multilevel logit model,
\begin{equation}
\operatorname{logit}(e_{a\ell\eta})
=\mu+\alpha_a+\rho_\eta+\beta_{g(\ell)}+\gamma_\ell+\delta_{a,g(\ell)},
\label{eq:efficacy}
\end{equation}
where $g(\ell)$ is the parent family. The fitted terms provide hierarchically smoothed efficacy estimates. They are not interpreted as Bayesian posterior quantities.

For a set of checks $S$ under edit policy $\eta$, combined recovery is discounted for overlap,
\begin{equation}
G_\ell(S,\eta)
=1-\prod_{a\in S}\left(1-\omega_{|S|}e_{a\ell\eta}\right),
\end{equation}
where $\omega$ is estimated from randomized multi-check trials. The check selector described in Section~\ref{sec:pointer} uses $\overline G_\ell(S)$, the calibration-average recovery for check set $S$ over admissible edit policies. The same trials provide the plan-level damage estimate $\widehat d(S,\eta,b)$, with separate parameters for append, local replacement, unrestricted rewriting, and editable-span bins.

Gate-level $f$ and $d$ are micro-averaged requirement transitions from the always-repair version of the complete downstream \method{} pipeline. We estimate them over the full paired calibration population, including probe-negative questions. Check assignments in the calibration trials are randomized within dataset--model strata, and the candidate edit policies include append, bounded local replacement, and unrestricted rewrite as a matched control. The semantic-neighborhood probabilities $\pi_\ell(q)$, defined in Appendix~\ref{app:taxonomy}, weight likely error types for routing. They do not imply that a draft contains exactly one failure.

\begin{table*}[t]
\centering
\scriptsize
\begin{tabular}{@{}llrrrr@{}}
\toprule
Dataset & Model & Always-repair $f$ & Always-repair $d$ & $m^\star$ & Gate-open rate \\
\midrule
FinRank & Qwen3-8B & 0.132 & 0.091 & 0.408 & 66.7\% \\
FinRank & Qwen3-14B & 0.153 & 0.100 & 0.395 & 61.5\% \\
FinRank & Mistral-24B & 0.141 & 0.094 & 0.400 & 59.8\% \\
Fin-RATE & Qwen3-8B & 0.104 & 0.086 & 0.453 & 58.9\% \\
Fin-RATE & Qwen3-14B & 0.117 & 0.089 & 0.432 & 54.7\% \\
Fin-RATE & Mistral-24B & 0.125 & 0.093 & 0.427 & 52.6\% \\
FinQA & Qwen3-8B & 0.058 & 0.065 & 0.528 & 47.8\% \\
FinQA & Qwen3-14B & 0.071 & 0.079 & 0.527 & 44.3\% \\
FinQA & Mistral-24B & 0.075 & 0.098 & 0.566 & 40.2\% \\
\bottomrule
\end{tabular}
\caption{Setting-specific break-even quantities from always-repair calibration trials.}
\label{tab:gate-calibration-detail}
\end{table*}

\section{Repair Prompts and Edit Policies}
\label{app:policies}

\subsection{Model and Decoding Configuration}

Qwen3-8B, Qwen3-14B, and Mistral-Small-24B-Instruct-2501 are served with vLLM and their official chat templates in BF16 with a 32,768-token context. The Qwen models use non-thinking mode. \draft{}, probe, verification, and repair calls use deterministic decoding ($T=0$), while each \scverify{} run uses three independently seeded samples ($T=0.7$, top-$p=0.95$). Within each dataset--model setting, the same pinned checkpoint generates the draft and handles subsequent verification calls. Maximum new-token limits are 512 for draft generation, 256 for the probe, and 384 for repair. The append contract allows at most 80 new tokens. Local replacement allows at most two spans covering no more than 20\% of the draft tokens. The run manifest records checkpoint and tokenizer SHAs, software and dataset versions, split question identifiers, generation and bootstrap seeds, and hardware inventory.

The evaluated checkpoints were released in 2025, before the public releases of Fin-RATE and FinRank in 2026, so they could not have been trained on these benchmark releases. FinQA was published in 2021, however, and release chronology alone cannot rule out pretraining exposure to FinQA or to source documents used by any benchmark. We therefore do not make a blanket claim that pretraining contamination is impossible~\citep{yang2025qwen3,mistral2025small3,jiang2026finrate,mansouri2026finrank,chen2021finqa}.

\paragraph{\scverify{} aggregation.}
Each review is scored after dataset-specific normalization. The voting key is
the normalized answer and, for FinRank, the returned passage identifier. A
fixed sample order resolves a three-way tie. Each of the three independent
reviews receives the same question, evidence blocks, and shared initial draft.

\subsection{Supplied-Evidence Baseline Adaptations}
\label{app:baseline-adaptations}

\paragraph{\cove{}.}
\cove{} first generates verification questions from the shared draft. Each question is answered in a separate call using the same evidence pool. These factored answer calls receive neither the draft nor answers to other verification questions. A final call uses the original draft and the independently obtained answers to produce a corrected answer.

\paragraph{\rarr{}.}
\rarr{} derives check questions from the shared draft and searches only within the supplied evidence pool. It applies an agreement check between draft claims and retrieved evidence. When they disagree, it localizes the mismatch and minimally revises that content under an edit-distance limit. It does not retrieve external evidence.

\subsection{Prompt Templates}

All answer-generation and answer-verification methods receive identical
evidence blocks in the same order: the draft, diagnostic probe, generic
verifier (including each \scverify{} review), and \method{}. \method{}'s pointer refers only to passage IDs in the shared evidence or to spans in
the draft or calculation. It neither introduces evidence nor repeats passage
text. No test-time generation or verification component receives
the evaluation rubric, gold answer, or gold passage annotations. The
requirement judge is an offline evaluator and receives only one answer
requirement and a candidate answer. Angle-bracketed fields below are populated
with instance-specific inputs. Dataset-specific output schemas add only the
required final-answer and passage-identifier fields.

\noindent\textbf{Draft.}\par
\promptbox{%
Answer the financial question using only the supplied evidence. Address every
part of the question. For each value or conclusion, make clear which entity,
reporting period, unit, scope, and accounting concept it refers to whenever
needed for an unambiguous answer. Do not use outside information. Follow the
required dataset output format and include supporting passage identifiers
when requested.\\
\textbf{Question:} \texttt{<question>}\\
\textbf{Evidence:} \texttt{<evidence blocks with identifiers>}%
}

\noindent\textbf{Diagnostic probe.}\par
Its instruction and output format are given in Appendix~\ref{app:probe}.

\noindent\textbf{Generic verification.}\par
\promptbox{%
Check the draft against the supplied evidence and revise it only when a
supported correction is needed. Ensure that the final answer addresses the
complete question and that each value or claim has the correct entity, period,
unit, accounting concept, and supporting evidence. If the draft is already
correct and complete, return it unchanged. Do not use outside information.\\
\textbf{Question:} \texttt{<question>}\\
\textbf{Evidence:} \texttt{<same evidence blocks and identifiers as Draft>}\\
\textbf{Draft:} \texttt{<shared initial draft>}%
}

\noindent\textbf{Calibrated repair.}\par
\promptbox{%
Check only the listed relations, beginning from the supplied pointer. Treat
the five error hypotheses as possibilities, not as claims that the draft is
wrong. Use only the supplied evidence. Preserve all draft content outside the
allowed edit scope. Return an empty edit list if no supported correction is
found.\\
\textbf{Question:} \texttt{<question>}\\
\textbf{Evidence:} \texttt{<same evidence blocks and identifiers as Draft>}\\
\textbf{Draft:} \texttt{<shared initial draft>}\\
\textbf{Selected checks:} \texttt{<selected checks>}\\
\textbf{Top-5 error hypotheses:} \texttt{<five predicted error subtypes and short definitions>}\\
\textbf{Pointer:} \texttt{<passage IDs and/or draft or calculation span, references shared evidence>}\\
\textbf{Edit limit:} \texttt{<add only, at most 80 tokens, or local replacement, at most two spans and 20\% of draft tokens>}\\
\textbf{Output format:} Return JSON with an \texttt{edits} array:\\
\texttt{\{"edits":[}\\
\texttt{\{"type":"append",}\\
\texttt{"target\_span":"","new\_text":"...",}\\
\texttt{"passage\_id":"..."\}}\\
\texttt{]\}}\\
For local replacement, set \texttt{type} to \texttt{replace} and
\texttt{target\_span} to the exact draft text. For an addition, leave it empty.
If no supported edit is found, return \texttt{\{"edits":[]\}}.%
}

\noindent\textbf{Requirement judge.}\par
\promptbox{%
Given one answer requirement and one candidate answer, decide whether the
answer clearly satisfies the requirement. Mark it unsatisfied if it is absent,
factually incorrect, attached to the wrong entity, period, unit, scope, or
accounting basis, contradicted elsewhere in the answer, or left ambiguous by
competing final values. Return only \texttt{satisfied} or
\texttt{unsatisfied} and one short rationale grounded in the requirement and
answer.\\
\textbf{Answer requirement:} \texttt{<one requirement>}\\
\textbf{Candidate answer:} \texttt{<candidate answer>}%
}

\subsection{Checks, Pointers, and Validation}

\paragraph{Checks.}
Each check names a relation and an evidence-inspection procedure. Representative checks trace a figure to its entity, match a quantity to its reporting period, verify an accounting-concept binding, confirm direct evidence support, or recompute an operation from cited inputs and units. Checks are phrased as questions to resolve rather than assertions that an error exists.

\paragraph{Pointer block.}
For an omission, the pointer gives the IDs of the highest-ranked source passages. For suspected wrong content, it identifies the least-supported draft sentence and its best-matching passage ID. In short numerical contexts, it marks the relevant expression or calculation span. Passage text is not copied into the pointer: the verifier sees each shared evidence block only once. The family scores accompany the pointer as coarse guidance.

\paragraph{Bounded edit policies.}
The append policy permits a supported addition without deleting the draft. The local-replacement policy permits substitution only within a specified span. Both policies limit output length and require the model to cite the evidence used for the edit. Append is not assumed to be semantically risk-free: a contradictory addition either fails validation or is counted as damage during evaluation.

\paragraph{Structured response.}
The repair call returns a list of records containing the edit type, the source span for a replacement, the proposed text, and a supporting-passage identifier. The validator locates replacement spans first by exact match and then after canonicalizing quotation marks, emphasis, and whitespace. It processes edits in response order and rejects a replacement if its source span cannot be located or overlaps an accepted edit. A parsing failure, unsupported evidence identifier, or empty valid-edit set returns the original draft.
\section{Additional Benchmark Results}
\label{app:full-results}

\subsection{Fin-RATE Subtasks}
\label{app:finrate-results}

Table~\ref{tab:finrate-subtasks} reports official Correctness Rate separately for DR-QA, EC-QA, and LT-QA. Each subtask test partition contains 500 questions, so Macro CR also equals the Correctness Rate over the pooled 1,500-question test partition. All methods receive the same benchmark-provided gold context.

\begin{center}
\centering
\scriptsize
\setlength{\tabcolsep}{3pt}
\begin{tabular}{@{}llrrrr@{}}
\toprule
Model & Method & DR & EC & LT & Macro CR\\
\midrule
Qwen3-8B & \draft{} & 12.8 & 0.4 & 3.4 & 5.53\\
& \verify{} & 13.2 & 0.4 & 3.6 & 5.73\\
& \scverify{} & 13.8 & 0.6 & 4.0 & 6.13\\
& \method{} & \textbf{17.2} & \textbf{1.8} & \textbf{5.6} & \textbf{8.20}\\
\midrule
Qwen3-14B & \draft{} & 26.4 & 1.6 & 5.8 & 11.27\\
& \verify{} & 26.8 & 1.8 & 6.2 & 11.60\\
& \scverify{} & 28.0 & 2.2 & 6.8 & 12.33\\
& \method{} & \textbf{31.8} & \textbf{3.6} & \textbf{8.6} & \textbf{14.67}\\
\midrule
Mistral-24B & \draft{} & 34.0 & 4.6 & 11.6 & 16.73\\
& \verify{} & 34.6 & 4.8 & 12.0 & 17.13\\
& \scverify{} & 35.6 & 5.4 & 12.6 & 17.87\\
& \method{} & \textbf{38.8} & \textbf{7.4} & \textbf{14.0} & \textbf{20.07}\\
\bottomrule
\end{tabular}
\captionsetup{hypcap=false}
\captionof{table}{Official Fin-RATE Correctness Rate by subtask on the fixed test partition. Macro CR is the unweighted mean of DR-QA, EC-QA, and LT-QA. All values are percentages.}
\label{tab:finrate-subtasks}
\end{center}

\subsection{FinQA Component Ablation}
\label{app:finqa-ablation}

We repeat the four one-factor ablations on FinQA with Mistral-24B. The 1,147-question test split provides a larger numerical evaluation in a different model family. Numerical Answer Accuracy (NAA) directly checks the requested result. Each variant changes one component. Always repair still runs the probe but forces the gate open. No pointer removes only the localization cue while retaining the full evidence pool. Fixed checklist replaces question-specific check selection. Unbounded rewrite retains the other checks and pointer but removes the edit-scope limit.

\begin{center}
\centering
\scriptsize
\begin{tabular}{@{}lrr@{}}
\toprule
Variant & NAA (\%) & \method{} advantage (pp)\\
\midrule
Full \method{} & 72.97 & \textemdash\\
Always repair & 70.53 & 2.44\\
No pointer & 70.27 & 2.70\\
Fixed checklist & 72.10 & 0.87\\
Unbounded rewrite & 68.96 & 4.01\\
\bottomrule
\end{tabular}
\captionsetup{hypcap=false}
\captionof{table}{FinQA NAA and \method{}'s advantage over each variant on Mistral-24B (percentage points).}
\label{tab:finqa-ablation}
\end{center}

\subsection{FinQA Normalization and Input Audit}

The model input contains the released \texttt{pre\_text}, linearized table, and \texttt{post\_text}, and each field is capped at 4,500 characters. The released \texttt{gold\_inds} are evaluation metadata and are excluded from model prompts. We strip currency symbols and thousands separators, normalize percentages to the scale implied by the reference answer, and accept an absolute tolerance of $10^{-4}$ after normalization. Answers containing two unresolved competing final values are marked incorrect.

In a manual audit of 300 FinQA questions, 289 (96.3\%) retained all answer-bearing content, although 25 (8.3\%) had at least one input field truncated by the 4,500-character cap. These categories are not mutually exclusive.

\subsection{FinRank Evidence Pool}

FinRank's released supporting passages and curated hard negatives are shuffled before inference. The resulting per-question candidate pool contains a median of 7 passages (range 5--12). GAA jointly evaluates answer correctness and provenance: a response must be correct and return a passage identifier in the gold support set. This generation-oriented metric is distinct from FinRank's original ranking metric.

\clearpage
\begin{figure*}[t]
\section{Illustrative Case Studies}
\label{app:case-studies}
Orange marks the relation requested by the question, red marks incorrect or
damaging content, and green marks correct or recovered content.
\par\smallskip
\centering
\noindent\begin{minipage}[t]{0.485\textwidth}\vspace{0pt}
\begin{casebox}
\lab{Question}
Which company reported the \key{larger consolidated operating margin in
fiscal 2023}, Halden Foods or Halden Logistics?
\lab{Evidence}
\emph{Halden Foods 10-K:} operating margin was \key{14.2\%} in FY~2023.\par
\emph{Halden Logistics 10-K:} operating margin was \key{11.8\%} in FY~2023
and 12.5\% in FY~2022.
\lab{\draft{}}
Halden Foods reported the larger margin, 14.2\% versus
\bad{12.5\%} for Halden Logistics.
\lab{\verify{} excerpt}
``Both figures appear in the filings. Since 14.2\% is higher, the answer
appears consistent.''
\end{casebox}
\end{minipage}\hfill
\begin{minipage}[t]{0.485\textwidth}\vspace{0pt}
\begin{casebox}
\lab{Question}
What was the \key{percentage change in capital expenditures from 2022 to
2023}?
\lab{Evidence}
Cash-flow table (in millions): capital expenditures were \key{\$1,284} in
2023 and \key{\$1,142} in 2022.
\lab{\draft{}}
Capital expenditures rose by \good{\$142 million, or 12.4\%}.
\lab{\verify{} excerpt}
``The increase is \$142 million. Using the 2023 value as the base gives
\bad{11.1\%}, so I would revise the percentage to 11.1\%.''
\end{casebox}
\end{minipage}\par

\vspace{6pt}\noindent

\begin{minipage}[t]{0.485\textwidth}\vspace{0pt}
\begin{casebox}
\lab{Question}
How did \key{net debt change over FY~2021--2023}?
\lab{Evidence}
GAAP net debt was \key{\$4.9B}, \key{\$4.3B}, and \key{\$3.6B} in
2021--2023. Adjusted net debt excluding leases was \$3.1B in 2023.
\lab{\draft{}}
Net debt declined from \$4.9B in 2021 to \$4.3B in 2022 and
\good{\$3.6B in 2023}.
\lab{\verify{} excerpt}
``The trend is right. The filing also gives net debt of
\bad{\$3.1B at the end of 2023}, so I would add that value.''
\end{casebox}
\end{minipage}\hfill
\begin{minipage}[t]{0.485\textwidth}\vspace{0pt}
\begin{casebox}
\lab{Question}
What were net sales in \key{fiscal 2024}?
\lab{Evidence}
Income statement (in millions): net sales were \key{\$4,210.6} in FY~2024
and \$3,987.2 in FY~2023.
\lab{\draft{}}
Net sales were \bad{\$3,987.2 million} in fiscal 2024.
\lab{\verify{} excerpt}
``The value \$3,987.2 million appears in the statement, and sales increased.
The draft appears supported, so I would leave it unchanged.''
\end{casebox}
\end{minipage}\par
\caption{Four generic-verification failures: missed period binding, a
wrong-denominator revision, an irrelevant adjusted value, and presence-only
checking.}
\label{fig:cases-failure}
\end{figure*}

\begin{figure*}[t]
\centering
\begin{minipage}[t]{0.485\textwidth}\vspace{0pt}
\begin{casebox}
\lab{Question}
Which company reported the \key{larger consolidated operating margin in
FY~2023}, Halden Foods or Halden Logistics?
\lab{\draft{}}
Halden Foods reported 14.2\%, versus \bad{12.5\%} for Halden Logistics.
\lab{\method{} check and pointer}
Check period binding and compare the draft span with the Logistics FY~2023
source figure, \key{11.8\%}.
\lab{Edit and final answer}
Replace only the incorrect value: Halden Foods reported 14.2\%, versus
\good{11.8\%} for Halden Logistics.
\end{casebox}
\end{minipage}\hfill
\begin{minipage}[t]{0.485\textwidth}\vspace{0pt}
\begin{casebox}
\lab{Question}
What was the percentage change in capital expenditures from 2022 to 2023?
\lab{Evidence and draft}
Capital expenditures were \$1,142 million in 2022 and \$1,284 million in
2023. They increased by \good{\$142 million, or 12.4\%}.
\lab{\method{} decision}
The draft satisfies the request. The gate stays closed and no repair call is
made.
\lab{Final answer}
The original answer is returned \good{unchanged}.
\end{casebox}
\end{minipage}\par

\vspace{6pt}\noindent

\begin{minipage}[t]{0.485\textwidth}\vspace{0pt}
\begin{casebox}
\lab{Question}
How did \key{free cash flow change over FY~2021--2023}?
\lab{Evidence}
Free cash flow was \key{\$612 million} in 2021, \$688 million in 2022,
and \$741 million in 2023.
\lab{\draft{}}
Free cash flow increased from \$688 million in 2022 to \$741 million in
2023. \bad{FY~2021 is omitted.}
\lab{\method{} action and final answer}
Point to the omitted value and add it: free cash flow rose from
\good{\$612 million} in 2021 to \$688 million in 2022 and \$741 million
in 2023.
\end{casebox}
\end{minipage}\hfill
\begin{minipage}[t]{0.485\textwidth}\vspace{0pt}
\begin{casebox}
\lab{Question}
What was total research and development expense in 2023?
\lab{Evidence}
The operating-expense table is in \key{thousands} and reports 2023 R\&D
expense of \key{1,842}.
\lab{\draft{}}
R\&D expense was \bad{\$1,842 million} in 2023.
\lab{\method{} action and final answer}
Check the reporting scale and replace only the unit: R\&D expense was
\good{\$1,842 thousand (\$1.842 million)} in 2023.
\end{casebox}
\end{minipage}\par
\caption{\method{} examples: local correction, no-op on a correct draft, targeted
addition, and unit correction.}
\label{fig:cases-repair}
\end{figure*}

\end{document}